\documentclass{ws-jmrr}
\usepackage[sort,compress]{cite}
\usepackage{bm}
\usepackage{color,soul}

\begin{document}

\catchline{0}{0}{2013}{}{}

\markboth{Sharon and Nisky}{What Can Spatiotemporal Characteristics of Movements in RAMIS Tell Us?}

\title{What Can Spatiotemporal Characteristics of Movements in RAMIS Tell Us?}

\author{Yarden Sharon and Ilana Nisky}

\address{Department of Biomedical Engineering, Ben-Gurion University of the Negev, Beer-Sheva, Israel\\
	Zlotowski Center for Neuroscience, Ben-Gurion University of the Negev, Beer-Sheva, Israel\\
	E-mail: nisky@bgu.ac.il}

\maketitle

\begin{abstract}
	Quantitative characterization of surgical movements can improve the quality of patient care by informing the development of new training protocols for surgeons, and the design and control of surgical robots. Here, we present a novel characterization of open and teleoperated suturing movements that is based on principles from computational motor control. We focus on the extensively-studied relationship between the speed of movement and its geometry. In three-dimensional movements, this relationship is defined by the one-sixth power law that relates between the speed, the curvature, and the torsion of movement trajectories. We fitted the parameters of the one-sixth power law to suturing movements of participants with different levels of surgical experience in open (using sensorized forceps) and teleoperated (using the da Vinci Research Kit / da Vinci Surgical System) conditions from two different datasets. We found that teleoperation significantly affected the parameters of the power law, and that there were large differences between different stages of movement. These results open a new avenue for studying the effect of teleoperation on the spatiotemporal characteristics of the movements of surgeons, and lay the foundation for the development of new algorithms for automatic segmentation of surgical tasks.

\end{abstract}


\begin{multicols}{2}
\section{Introduction}

Surgery is a complex motor task that requires planning and execution of accurate movements, and typically, surgeons need many years of training and many cases to master surgery \cite{RefWorks:71,RefWorks:138}. Computational motor control studies aim to develop computational models that explain how the brain controls the movements of our body, adapts, and learns \cite{RefWorks:153, RefWorks:113,RefWorks:142,RefWorks:143,RefWorks:144,RefWorks:145}. These models quantify various regularities in human motion. For example, planar point-to-point movements are characterized by straight paths \cite{RefWorks:146} and bell-shaped velocity profiles \cite{RefWorks:39}, and planar curved movements are characterized by the two-thirds power law that relates the speed of motion to its geometry \cite{RefWorks:15}. In three-dimensional movements, this spatiotemporal relationship is defined by the one-sixth power law that relates between the speed, the curvature, and the torsion of movement trajectories \cite{RefWorks:95,RefWorks:42}. In this paper, we employ the one-sixth power law to quantitatively characterize surgical movements.  

Today, a variety of devices are available that can facilitate motion tracking in surgery, such as external tracking systems  \cite{RefWorks:34,RefWorks:75}, virtual reality simulators \cite{RefWorks:35,RefWorks:79}, and robot-assisted surgery systems \cite{RefWorks:119, RefWorks:33}. Therefore, models and theories of computational motor control can be combined with data for understanding how surgeons control the movements of their hands and instruments \cite{RefWorks:137}. Such understanding may help to develop effective training protocols for new surgeons, to improve the design and control of teleoperated surgical robots, and even serve as inspiration for the development of autonomous or semi-autonomous surgical systems. 
 
 It is known that the skill of a surgeon influences the outcome of surgical procedures \cite{RefWorks:27}. Therefore, great efforts are invested in improving the acquisition and evaluation of surgical skill. Today, surgical skill assessment is mostly subjective, and it is done by surgeons who observe and rate other surgeons using global rating scales such as GOALS and OSATS \cite{Refworks:147,RefWorks:31}. The disadvantages of such subjective assessment are high variability between different observers, and the lack of information available to the observers. Thus, it is important to develop objective metrics for surgical skill that will describe surgical performance in many details. Examples of such objective measures that are used today are task completion time \cite{RefWorks:2,RefWorks:62,RefWorks:63} and path length \cite{RefWorks:66,RefWorks:61,RefWorks:56}. However, these metrics are not sufficient for an accurate skill assessment. Other recent metrics are based on movement smoothness \cite{RefWorks:80,RefWorks:148}, energy  \cite{RefWorks:149,RefWorks:150}, contact forces and robot arm acceleration \cite{RefWorks:151}, and others. We suggest that quantifying the spatiotemporal characteristics of movements may be useful in the development of new objective metrics for surgical skill.

Quantitative characterization of surgical movements may also contribute to the design of surgical robots. In teleoperated robot-assisted minimally-invasive surgery (RAMIS), surgeons use robotic manipulators to control the movements of instruments that are inserted into the body of the patient via small incisions \cite{RefWorks:5}. Teleoperated robot-assisted surgery offers many advantages over open surgery \cite{RefWorks:50,RefWorks:72}. However, it also has drawbacks, such as the absence of touch information \cite{RefWorks:73}, and the effects of the dynamics of the robot on the movement of the surgeon \cite{RefWorks:8,RefWorks:36}. As a first step towards improving the design of surgical robots, and making RAMIS teleoperation more natural, it is important to quantify how teleoperation affects the performance of surgical tasks compared to open surgery. Therefore, in this paper, we investigate how teleoperation affects the spatiotemporal characteristics of movements.

Segmentation of complex surgical tasks can also benefit from motor control models and theories. It is very easy to record surgical simulation or RAMIS data for analysis, but the resulting datasets are very long, and therefore, may be difficult to parse. To analyze surgical movement data effectively, it is necessary to divide the complex surgical movements into simpler segments. Thus, developing algorithms for automatic segmentation of surgical tasks can contribute to surgical skill assessment \cite{RefWorks:82} and other related fields. Previous studies used various machine learning algorithms for surgical task segmentation \cite{RefWorks:85,RefWorks:87,RefWorks:105, RefWorks:170}. Segmentation of complex movements into smaller segments was also studied in human motor control \cite{RefWorks:19}. We suggest that a similar approach may be also useful for the segmentation of complex surgical tasks. Such segmentation will highlight segments that are distinct in terms of their movement coordination rather than contextual information.

With these three important applications in mind, in this study, we analyzed teleoperated and open suturing movements of participants with different levels of surgical experience. We used two different datasets that focused on different aspects of surgical suturing. The first dataset was collected during teleoperated (using the da Vinci Research kit -- dVRK \cite{RefWorks:33}) and open suturing movements \cite{RefWorks:11}. The second dataset is teleoperated suturing movements (using the da Vinci Surgical System -- dVSS \cite{RefWorks:119}, Intuitive Surgical, Inc., Sunnyvale, CA) from the JHU-ISI Gesture and Skill Assessment Working Set (JIGSAWS) \cite{RefWorks:112} -- a surgical activity dataset for human motion modeling. We used the one-sixth power law to investigate how expertise level, teleoperation (open suturing compared to robot-assisted suturing), and movement segment (within the suturing task), affect the coordination between movement speed, curvature, and torsion.

\section{One-Sixth Power Law}
The relationship between the speed of movement and its geometry was studied extensively. The two-third power law describes an inverse relationship between speed and curvature in planar drawing movements, and it is often stipulated as follows \cite{RefWorks:15}:
\begin{equation}
v =\alpha\kappa^{-\frac{1}{3}},
\end{equation}
where $v$ is the movement speed, $\alpha$ is a constant, called the speed gain factor, and $\kappa$ is the curvature of the path, which measures the local deviation from straightness \cite{RefWorks:95}. Note that the origin of this power law's name is another form of the same relation, using the angular speed $\omega$ ($\omega=\alpha\kappa^{2/3}$).

Movement in accordance with the two-third power law is equivalent to motion at constant planar affine speed \cite{RefWorks:97,RefWorks:98}. Many studies documented the existence of the two-third power law in a variety of movements, including drawing movements \cite{RefWorks:17}, eye movements \cite{RefWorks:21} , human locomotor trajectories \cite{RefWorks:20}, and even imagined trajectories \cite{RefWorks:152}. However, it was found that this power law does not adequately describe three dimensional movements \cite{RefWorks:96}. Instead, using a generalization of motion at constant affine speed from two to three dimensions, a three dimensional power law was proposed -- the one-sixth power law \cite{RefWorks:95,RefWorks:42}. This power law describes the relation between the speed of the movement, its curvature, and its torsion \cite{RefWorks:95}:  
\begin{equation}
v =\alpha\kappa^{-\frac{1}{3}}|\tau|^{-\frac{1}{6}},
\end{equation}
where $\alpha$  is constant and called the speed gain factor, $\kappa$ is the curvature, and $\tau$ is the torsion of the path, which measures the local deviation from planarity \cite{RefWorks:95}.

One approach to investigate the relationship between speed, curvature, and torsion is to use regression for fitting the exponents of the power law together with the speed gain factor, i.e.:
\begin{equation}
v =\alpha\kappa^\beta|\tau|^\gamma.
\label{eq:exponents}
\end{equation}
A study that used this approach in the analysis of shape tracing found that the means of the values of the exponents $\beta$ and $\gamma$ were different between the different shapes \cite{RefWorks:42}. We adopted this approach, and employed this analysis to study the relationship between speed, curvature, and torsion in suturing movements. We assumed that the values of the exponents will be similar to the exponents of the one-sixth power law. In addition, we expected that expertise level, teleoperation condition, and movement segment will affect the values of the power law parameters.

The first effect that we examined is expertise level. Previous studies used the speed gain factor $\alpha$ from the two-third power law to predict surgical skill \cite{RefWorks:115}. In addition, for drawing movements, it was previously found that the exponent of the two-third power law depends on age. For children, the exponent differs from --1/3, and it converges to the typical adult value with age \cite{RefWorks:17}. However, it is not clear whether this change in the exponent is caused by the difference between the motor systems of children and adults, or by the learning of new motor tasks. In our study, some of the participants have no surgical experience, and therefore, the suturing task is completely novel to them. In contrast, some of the participants are experienced surgeons who are already familiar with the task. Therefore, we hypothesized that the parameters of the one-sixth power law in suturing movements may reveal differences between participants with different levels of surgical expertise.

The second effect we investigated is teleoperation. The dynamics of the surgical manipulators and the teleoperation controllers impose challenges and constraints on the motor control system of the surgeon; these factors affect user movements \cite{RefWorks:36, RefWorks:8}. In addition, a recent study showed that the exponent of the two-thirds power law (Eq. 1) in water is significantly different than in air \cite{RefWorks:116}, suggesting that dynamic constrains affect the relationship between the speed and curvature in planar movements. Based on this result, we assumed that the teleoperation-induced constraints would be reflected as differences in the parameters of the one-sixth power law between the teleoperated and the open condition. 

The last effect that we investigated is differences between different movement segments. Studies showed that movements in different shapes resulted in different values of exponents  \cite{RefWorks:42,RefWorks:117}. Based on these results, we hypothesized that different segments of the suturing movement will produce different values of exponents when fitting the data to the power law model. In addition, previous studies showed that the speed gain factor of the two-third power law varies between different segments of movement, and may also help in movement segmentation. For example, this approach was used for segmentation of the complex movements in sign language \cite{RefWorks:99}, and for segmentation of surgical tasks \cite{RefWorks:120,RefWorks:121}. Therefore, we expected that the values of the speed gain factor of the one-sixth power law may also be used for segmentation of the suturing movements.

\section{Methods}
In this study, we analyzed two datasets. The first dataset, called here dVRK-Open, is data that was collected in a previous study  during teleoperated (using the da Vinci Research kit -- dVRK \cite{RefWorks:33}) and open suturing movements \cite{RefWorks:11}. The second dataset, called JIGSAWS, is teleoperated suturing movements (using the da Vinci Surgical System -- dVSS \cite{RefWorks:119}, Intuitive Surgical, Inc., Sunnyvale, CA) from the JHU-ISI Gesture and Skill Assessment Working Set \cite{RefWorks:112} -- a surgical activity dataset for human motion modeling. In the following subsections, we first describe each dataset with its corresponding data analysis (subsections 3.1 and 3.2), then we describe the fitting of the one-sixth power law (subsection 3.3), and we finish with a summary of the statistical analysis (subsection 3.4).

\subsection{Dataset 1: dVRK-Open}
Full details of the experimental setup and procedures for this dataset are reported in \cite{RefWorks:11}, but for completeness, we present the most important information below. 

\subsubsection{da Vinci Research Kit Setup}
The dVRK's setup that was used during the experiment consisted of a pair of Master Tool Manipulators (MTMs), a pair of Patient Side Manipulators (PSMs), four manipulator interface boards, a high resolution stereo viewer, and a foot-pedal tray. Two large needle-drivers were used as PSM instruments. All the components were mounted on a custom-designed extruded aluminum structure. The electronics and firmware of the interface boards were based on a custom IEEE-1394 FPGA board and quad linear amplifier \cite{RefWorks:33}. The MTM and PSM electronics were all connected via firewire connectors to a single computer with an Intel Core i7 4960X processor.

The participants watched a 3D view of the task scene. The visual scene acquired using a pair of Flea 3 cameras (Point Grey, Richmond, BC) with 16 mm f1.8 compact instrumentation lenses (Edmund Optics, Barrigngton, NJ) that was mounted on a custom designed fixture. The position and orientation of the camera were manually adjusted to achieve the best view of the task board. 

The control of the dVRK was based on position exchange with PD controllers. The joint angles were used to calculate the Cartesian positions and the orientations of the MTM and PSM tooltips via forward kinematics. The velocities were calculated using numerical differentiation and filtering with  a 2\textsuperscript{nd} order Butterworth low-pass filter with a 20 Hz cutoff. To control the PSM, the position and velocity of the MTM were down-scaled by factor of 3 to mimic the 'fine' movement scaling mode of the clinical da Vinci system. The orientation was not scaled. Similarly to the clinical da Vinci, there was no force feedback, and there was a small torque feedback on the orientation degrees of freedom to help users avoid large misalignment in tool orientation between PSM and MTM as a result of joint limits or singular configurations.

\subsubsection{Experimental Procedures}

 Sixteen participants took part in the experiment that was approved by the Stanford University Institutional Review Board, after giving informed consent. The participants were six experienced surgeons (five urologists with more than 120 robotic cases, and one general surgeon with more than 150 robotic cases, self reported), and ten non-medical participants (engineering graduate students). One non-medical participant had extensive experience with the experimental setup, and therefore, was removed from the analysis. In addition, during the experiment of one of the experienced participants, there were problems with the data recording, and this participant was also removed from the analysis.
 
  The participants performed teleoperated and open unimanual suturing. The order of the two sessions (teleoperated and open) was balanced across participants. In the teleoperated session, the participants performed the task using the dVRK. They sat in front of the master console, and the task-board was mounted on the patient-side table such that its position was fixed relative to the cameras (Fig.~\ref{fig1}(a)). Prior to each experiment, the master console ergonomics was adjusted so that the posture of the participant was comfortable.
  
   In the open session, the participants used a standard surgical needle-driver. To provide similar context to the teleoperated session, the participants also sat in front of the dVRK. A similar task-board was mounted on the armrest of the dVRK (Fig.~\ref{fig1}(b)). Two magnetic pose trackers (trakSTAR, Ascension Technology Corporation, Shelburne, VT) were mounted on the shafts of the needle-driver, and their positions were recorded. To prevent signal distortion, the tracker was separated from the metal body of the driver by 2 cm.

   The participants watched an introduction video before each session (teleoperated or open).
   The video contained explanations about bimanual needle handling technique, unimanual suturing, and acceptable correction movements.
    
   The task board consisted of four marks (Fig.~\ref{fig2}(a.III)): start (\textit{s}), insertion (\textit{i}), exit (\textit{e}), and finish (\textit{f}). The trial started with a bimanual adjustment of the needle in the right needle-driver in a configuration that is appropriate for driving the needle via the tissue. This adjustment was performed using the right and left needle-drivers in the teleoperated condition, and the needle-driver and the fingers of the left hand in the open condition. Then, participants placed the tip of the needle at start target (\textit{s}), and in the teleoperated condition, pressed the left foot-paddle to indicate the beginning of the task sequence. In the open condition, they
   pressed the left button of a computer mouse that was placed near the left hand of the participant on the armrest instead of the foot paddle.
   
    A single suturing trial included four segments (Fig.~\ref{fig2}(a)): (I) transport -- reaching with the head of the needle from \textit{s} to \textit{i}, (II) insertion -- driving the needle through the artificial tissue until its tip exits at \textit{e}, (III) catching -- opening the gripper and catching the tip of the needle, and (IV) extraction -- pulling the needle and reaching to \textit{f} with its tail. The trial ended when the tail of the needle was placed at the end target, and left foot-paddle or mouse-button were pressed to indicate trial end. In each session (teleoperated and open) the participants performed 80 trials, with breaks every 10 trials. 
    
During the experiment, some of the trials were not performed according the instructions, and some were not recorded properly. These trials were removed from the analysis. Among teleoperated sessions, 27 out of the 1120 trials of all the participants were removed, and in the open sessions, 51 out of the 1120 trials of all the participants were removed.

\subsubsection{Preprocessing}
In this study, we analyzed the Cartesian position of the needle-driver. In the teleoperated session, we used the\begin{figurehere}
	\begin{center}
		\centerline{\includegraphics{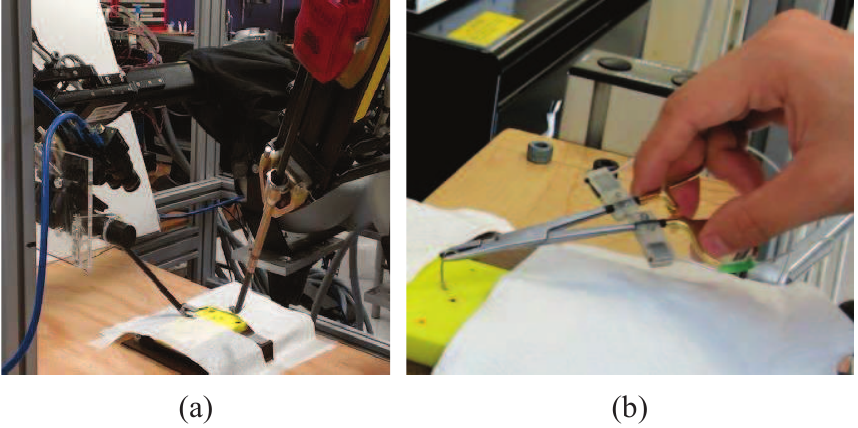}}
		\caption{dVRK-Open experimental setups in the teleoperated and open suturing. (a) The task board and the instruments on the patient-side table in the teleoperated session. (b) A surgical needle-driver with magnetic trackers in the open session.}
		\label{fig1}
	\end{center}
\end{figurehere} position of the right PSM, which was recorded at  2 kHz. In the open session, the positions of the two magnetic pose trackers were recorded at 120 Hz. We calculated the position of the driver's endpoint using a calibration dataset. Later, we interpolated and down-sampled all the positions data (from both sessions) to 100 Hz using piecewise cubic Hermite interpolating polynomial (PCHIP).

\subsubsection{Segmentation}
We built a segmentation algorithm that divided the movement into its four stages (transport, insertion, catching and extraction). The algorithm was based on the movement's trajectory and the opening angle of the needle-driver, full details of the segmentation process are reported in \cite{RefWorks:171}. Note that this segmentation algorithm was not based on the one-sixth power law. The third and the last segments were highly affected by whether the participants followed the instructions. They had substantial strategical variability, and therefore, our segmentation algorithm could not separate these two segments sufficiently. Hence, we focused only on the first and second segments.

\subsection{Dataset 2: JIGSAWS}
In the second part of our study, we used the JHU-ISI Gesture and Skill Assessment Working Set (JIGSAWS) \cite{RefWorks:112}. This dataset was collected using the da Vinci Surgical System (dVSS), and consists of kinematic and video recordings of elementary surgical tasks on a bench-top model. In this study, we used the kinematic data that was recorded during the suturing task. These data include recordings from eight surgeons with different robotic surgical experience (four surgeons with less than 10 hours,  two surgeons with 10--100 hours, and two surgeons with more than 100 hours, self reported).

\begin{figurehere}
	\begin{center}
		\centerline{\includegraphics{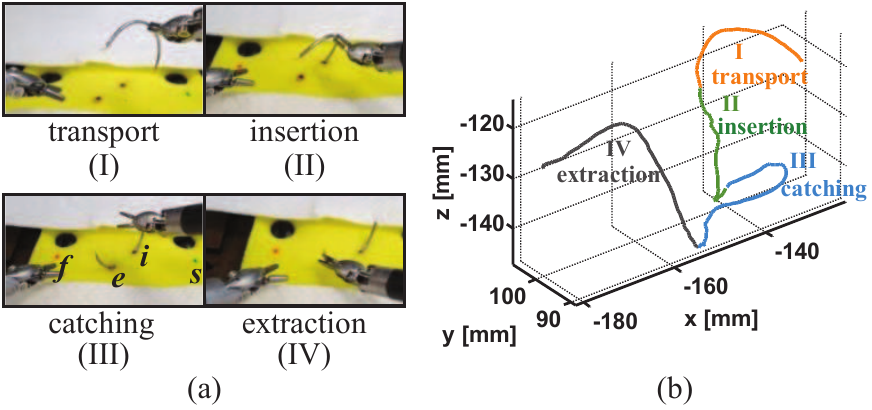}}
		\caption{dVRK-Open dataset task segments. (a) The task board and task segments: (I) transport, (II) insertion, (III) catching, (IV) extraction. (b) An example of the path of the right instrument. The numbers and the different colors indicate the four segments of the task.}
		\label{fig2}
	\end{center}
\end{figurehere}

 In each trial of the suturing task, the surgeon piked up a needle, reached the incision mark and passed the needle via the mock tissue between two marked targets. After the needle pass, the surgeon used the left tool to extract the needle from the tissue, and passed it to the right tool. Then, The surgeon repeated the needle pass three more times. While performing the task, the surgeon was not allowed to move the camera or apply the clutch. Each surgeon repeat the suturing task five times. The JIGSAWS includes 39 trials of the suturing task.
 In our analysis we used the Cartesian position of the PSMs that was recorded at 30 Hz.

\subsubsection{Segmentation}
 The JIGSAWS includes a manual segmentation of each movement into surgical gestures \cite{RefWorks:112}. This segmentation was done by an individual that watched the video and in consultation with a surgeon. We used this segmentation in our analysis. In this paper, we will refer to each surgical gesture as a segment. In the suturing task, there were ten segments. However, some of the segments were not observed in all the trials, and some of the segments included movement of both tools (right and left). Therefore, we chose to analyze only four segments (Fig.~\ref{fig3}): (G2) positioning -- positioning  needle, (G3) pushing -- pushing needle through tissue, (G6) pulling -- pulling the suture with left hand, and (G4) transferring -- transferring needle from left to right.

\subsubsection{Surgical Skill Annotation}
The JIGSAWS dataset also includes a manual annotation of surgical technical skill \cite{RefWorks:112}. Each trial in the JIGSAWS has a global rating score (GRS) that was assigned by an experienced surgeon who watched the videos. To assign the GRS, the surgeon used a modified objective structured assessments of technical skills (OSATS) method \cite{RefWorks:31}. The GRS 
\begin{figurehere}
	\begin{center}
		\centerline{\includegraphics{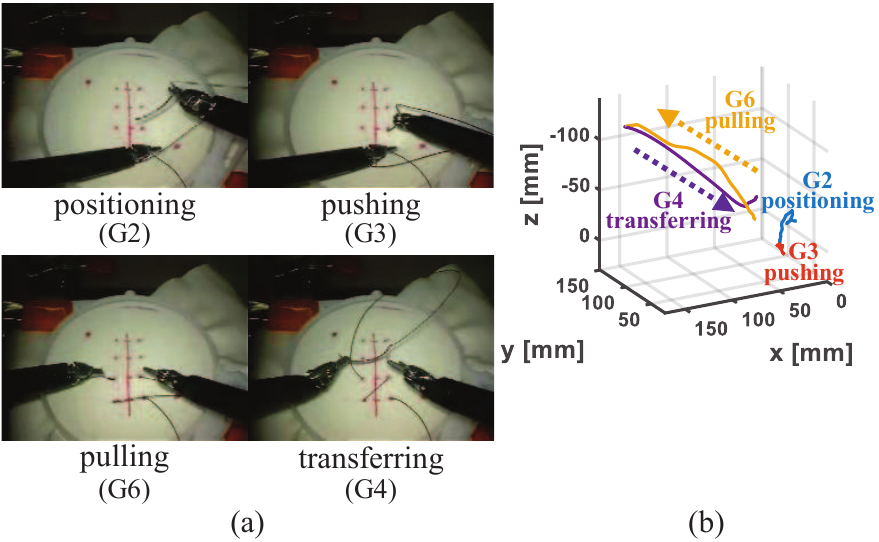}}
		\caption{JIGSAWS dataset task segments \cite{RefWorks:112}. (a) The task workspace and an example of the task segments: (G2) positioning, (G3) pushing, (G6) pulling, (G4) transferring. (b) An example of the path of the instruments from one suturing. The numbers and the different colors indicate the four segments of the task.}
		\label{fig3}
	\end{center}
\end{figurehere}
ranges from 5 (low technical skill) to 30 (high technical skill). For more information on the components of the GRS, see \cite{RefWorks:112}.
\subsection{Fitting the one-sixth power law}
 In this subsection, we will describe the procedure of extracting the parameters of the one-sixth power law from the data. First, We filtered the position data offline at 6 Hz with a 2\textsuperscript{nd} order zero lag Butterworth filter (Matlab's \texttt{filtfilt()}). Then, we used the filtered position $\bm{x}(t)=[x(t),y(t),z(t)]^T$ to calculate the first, second, and third time derivatives of the position ($\dot{\bm{x}}(t)$,    $\ddot{\bm{x}}(t)$, and   $\dddot{\bm{x}}(t)$, respectively) using numerical differentiation. We filtered each derivative before calculating the next order derivative. For the derivatives, we used a 2\textsuperscript{nd} order zero lag low-pass Butterworth filter with a cutoff at 10 Hz.

To fit the speed gain factor $\alpha$ and the exponents $\beta$ and $\gamma$, we first calculated the curvature $\kappa$ and the torsion $\tau$ for each sampling point. The curvature of the path was calculated as in \cite{RefWorks:96}: 
\begin{equation}
\kappa(t)=\sqrt{\frac{||\dot{\bm{x}}(t)||^2   ||\ddot{\bm{x}}(t)||^2  -  \left(\dot{\bm{x}}(t)^T\ddot{\bm{x}}(t)\right)^2  }
	{||\dot{\bm{x}}(t)||^6}},
\end{equation}
and the torsion was calculated as in \cite{RefWorks:42}:
\begin{equation}
\tau(t)=\frac{|\dot{\bm{x}}(t),    \ddot{\bm{x}}(t),   \dddot{\bm{x}}(t)|}{||\dot{\bm{x}}(t)\times\ddot{\bm{x}}(t) ||^2},
\end{equation}
where $|\bm{u},\bm{v},\bm{w}|$ denotes the scalar triple product of three vectors $\bm{u},\bm{v},\bm{w}$:  $|\bm{u},\bm{v},\bm{w}|=\bm{u}\bullet(\bm{v}\times\bm{w})$, the operator $\bullet$  denotes a dot product, and the operator $\times$ denotes cross product. We then took the log of both sides of Eq.~(\ref{eq:exponents}) to get:
\begin{equation}
\log{v}=\log{(\alpha)}+\beta\log{(\kappa)}+\gamma\log{(|\tau|)}
\label{eq:log}
\end{equation}
Then, for each segment in each trial, using the $v$, $\kappa$, and $\tau$ that were calculated from the sampled data, we fitted a linear regression model to find the values of the three parameters: the logarithm of the speed gain factor ($\log{(\alpha)}$), and the exponents $\beta$ and $\gamma$.

In some of the movements, there were small parts in which the movement was planar. In these parts, it is not possible to fit the one-sixth power law. To deal with this problem we used only sampling points in which the absolute value of the torsion was higher than  2 $m^{-1}$, this threshold was used in \cite{RefWorks:42}. In total, 0.3\% of the sampling points were below the threshold.

\subsection{Statistical Analysis} 
Each of the datasets was collected using different experimental procedures, and each had its own conditions and features. For example, only the dVRK-Open had an open condition, and only the JIGSAWS had GRS evaluation. Therefore,  for each dataset, we performed a different analysis, and therefore, we describe the different statistical analyses separately for each dataset. 
 
The statistical tests were performed using the Matlab Statistics Toolbox. In all the tests, statistical significance was determined at the 0.05 threshold. We used the Lilliefors test for the assumption of normality. Where needed, we used Mauchly's test for the assumption of sphericity, and none of our models violated sphericity. We used the Bonferroni correction for post-hoc comparisons, the Bonfferoni-corrected p values are denoted as $p_B$.
 
\subsubsection{dVRK-Open}
 We wanted to evaluate how expertise, teleoperation, and movement segment affect each of the power law parameters ($\alpha$,$\beta$,$\gamma$). For each of these parameters, separately for each participant,  segment, and teleoperation condition, we calculated the average value over the entire 80 trials. Our experimental design was mixed -- each participant belonged only to one expertise group (experienced surgeon or non-medical user), but performed all the segments in both conditions. Therefore, we fitted a 3-way mixed-model ANOVA with the average value across trials of the parameter as the dependent variable. The independent variables were: expertise (experienced surgeon/non-medical user, between participants), teleoperation condition (teleoperated/open, within participants), segment (I. transport/II. insertion, within participants), and their first and second order interactions.
 \subsubsection{JIGSAWS} 
For this dataset, we performed several statistical analyses. We first tested whether there is a correlation between the parameters of the power law ($\alpha$,$\beta$,$\gamma$) and surgical expertise, as measured by the global rating score (GRS). To examine the existence of such correlation, we used regression analysis. % We first calculated for each trial an average value of the power law parameter. 
Each participant performed five trials that consisted of several suturing movements, and hence, each trial had several repetitions of the different segments. We first calculated the average values of the parameter for the four segments in the trial. Then, using the average values of the four segments, we calculated one average value for each of the five trials. Then, we fitted regression models to this average as a function of the trial's GRS.

When fitting such regression models, two questions can be answered: the first one is whether there is a global effect of GRS on the power law parameters. This effect would be mainly affected by the difference in expertise between the surgeons. To answer this question, we fitted a linear regression model with the trial's average value of the parameter as the dependent variable and the trial's GRS as the independent variable. The second question, is whether within the different trials of each surgeon, there is a finer effect of the GRS on the power law parameters. To answer this question, we fitted a 1-way repeated-measures regression model with the trial's average value of the parameter as the dependent variable, and the trial's GRS as the independent variable. This model still fits a single slope for all the participants, but a different intercept is fitted to the data of each participant. We chose not to include in this analysis an interaction term between participant and GRS because of the limited number of data points for each participant.

% \subparagraph{Segments}
We were also  interested in the question how movement segment affects the parameters of the power law. Therefore, for each parameter ($\alpha$,$\beta$,$\gamma$), we calculated the average value across all the repetitions of each segment for each participant. In other words, for each participant, we calculated four values for each parameter. Then, we fitted a 1-way repeated-measures ANOVA model with the average value of the parameter as the dependent variable. The independent variable was segment (G2. positioning / G3. pushing / G6. pulling/ G4. transferring, within participant).

\section{Results}
Overall, the suturing movements from both datasets are described very well with the one-sixth power law. This is depicted in Fig.~\ref{fig4} that presents data from one segment that was analyzed in our study. Fig.~\ref{fig4}(a) describes the speed $v$, curvature $\kappa$, and torsion $\tau$ of the movement as a function of time. This graph demonstrates the inverse relation between the speed and the curvature, and the weaker inverse relation between the speed and the torsion. For example, the gray dashed lines in the plot highlight the local minima of the speed that are located in the same locations as the local maxima of the curvature. Similar effects happen also
\begin{figurehere}
	\begin{center}
		\centerline{\includegraphics{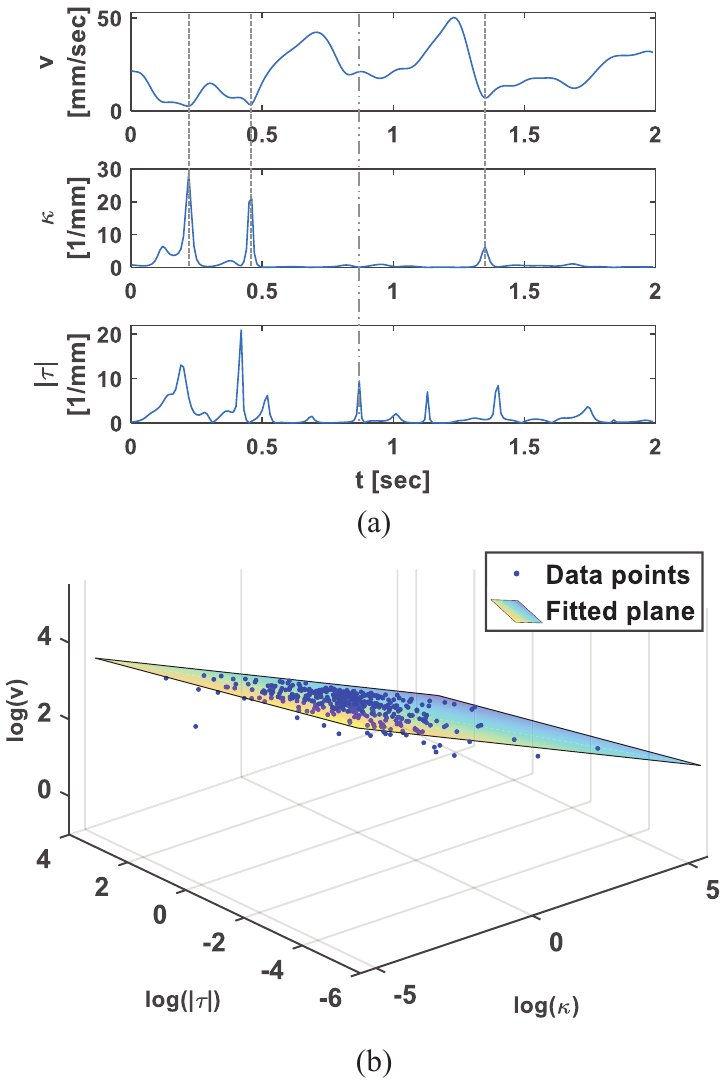}}
		\caption{An example from the data. (a) The speed $v$, curvature $\kappa$, and torsion $\tau$ of the movement as a function of time. (b) The relationship between $\log{(v)}$, $\log{(\kappa)}$, and $\log{(|\tau|)}$}
		\label{fig4}
	\end{center}
\end{figurehere}
   in the local maxima of the torsion, but they are masked in this plot by the effect of the curvature that has a larger power function. In addition, Fig.~\ref{fig4}(b)  illustrates the relationship between $\log{(v)}$, $\log{(\kappa)}$, and $\log{(|\tau|)}$. The participant's movement in this example is in accordance with the power law, as evident in the homogeneous spread of the data points around the fitted plane. This example is representative of both datasets, and is in agreement with the quantitative analysis that is summarized in Table 1. The overall goodness of fit of the movements to the power law is high ($R^2>0.82$), and the average values of the curvature exponent $\beta$ and the torsion exponent $\gamma$ are close to the theoretical values. In addition, the values of $\beta$ and $\gamma$ for the dVRK and the dVSS are similar, and slightly different from the open data.

\subsection{dVRK-Open} 
 The first effect that we investigated was expertise level. We found that the effect of expertise was statistically significant only for the speed gain factor $\alpha$ -- it was larger for the experienced surgeons compared to non-medical users 
\begin{tablehere}
 	\tbl{Fitting the One-Sixth Power Law for both datasets\label{tab1}}
 	{\begin{tabular}{@{}lccr@{}}
 			\toprule
 			& $\beta$      & $\gamma$     & R\textsuperscript{2}      \\ \hline
 			Open & -0.386$\pm$0.077 & -0.192$\pm$0.068 & 0.842$\pm$0.109 \\
 			(dVRK-Open)   &              &              &             \\\hline
 			dVRK & -0.402$\pm$0.073 & -0.197$\pm$0.053 & 0.827$\pm$0.101 \\
 			(dVRK-Open)   &              &              &             \\\hline
 			dVSS & -0.423$\pm$0.056 & -0.199$\pm$0.041 & 0.870$\pm$0.080 \\
 			(JIGSAWS)   &              &              &            \\
 			
 			\botrule
 	\end{tabular}}
 	\raggedright{All numerical values are mean$\pm 1$ standard deviation.}
 \end{tablehere}
 (Fig.~\ref{fig5}(a)). In addition, there was a statistically significant interaction between expertise and teleoperation condition. A post-hoc analysis revealed that the difference between $\alpha$ values of experienced surgeons and non-medical users in the open condition was bigger than in the teleoperated condition ($t_{12}=2.703$, $p_B=0.019$, and $t_{12}= 2.443$, $p_B=0.031$, respectively). These results suggest that the speed gain factor $\alpha$ can be useful for measuring surgical expertise level.

The second effect was teleoperation condition. We found that the speed gain factor $\alpha$ and the curvature exponent $\beta$ were both statistically-significantly smaller in the teleoperated compared to the open conditions (Fig.~\ref{fig5}(a-b)). For the torsion exponent $\gamma$, this difference was not statistically significant, but the interaction between teleoperation condition and segment was statistically significant (Fig.~\ref{fig5}(c)). A post-hoc analysis revealed that in segment I (transport), the value of $\gamma$ in the teleoperated condition was significantly higher than in the open condition ($t_{12}=2.237$, $p_B=0.045$). However, in segment II (insertion) the value of $\gamma$ in the teleoperated condition was significantly lower than in the open condition ($t_{12}=5.311$, $p_B<0.001$). These results show that teleoperation affected the movements of the user, and that the direction of this effect depends on the segment of movement.

The last effect for this dataset was movement segment. We found that there were statistically significant differences between the two segments for all the three parameters -- $\alpha$, $\beta$, and $\gamma$ (Fig.~\ref{fig5}(a-c)). In addition, for the speed gain factor $\alpha$, there was statistically significant interaction between segment and teleoperation condition. A post-hoc analysis revealed that the difference between the segments in the open condition was larger than the difference in the teleoperated condition ($t_{12}=4.346$, $p_B<0.001$, and $t_{12}=2.678$, $p_B=0.020$, respectively). These results suggest that the values of the power-law exponents and the speed gain factor may be used for building a segmentation algorithm for complex surgical tasks.

\subsection{JIGSAWS} 
To evaluate the effect of expertise on the parameters of the power law, we performed regression analysis of the parameters as a function of the GRS. The results of this analysis are depicted in Fig.~\ref{fig6}. The data of each surgeon is denoted
\begin{tablehere}
	\tbl{Statistical Analysis Summary -- dVRK-Open 	\label{tab2}}
	{\begin{tabular}{@{}lcccr@{}}
			\toprule
			Factor & Stat   & $\alpha$                & $\beta$                 & $\gamma$                \\  \colrule
			Expertise                       & $F_{1,12}$ & 7.435                   & 4.141                   & 0.022                   \\
		                        & $p$ & \textbf{0.018}          & 0.065                   & 0.884                   \\ \hline
			Teleoperation                   &$F_{1,12}$  & 301.355                 & 8.514                   & 0.041                   \\
		                       & $p$ & \textbf{\textless0.001} & \textbf{0.013}          & 0.842                   \\ \hline
			Segment                         &$F_{1,12}$  & 14.188                  & 47.68                   & 223.254                 \\
		                      & $p$ & \textbf{0.003}          & \textbf{\textless0.001} & \textbf{\textless0.001} \\ \hline
			Expertise*Teleoperation         & $F_{1,12}$  & 5.707                   & 0.211                   & 0.003                   \\
		                       & $p$$p$ & \textbf{0.034}          & 0.654                   & 0.959                   \\ \hline
			Expertise*Segment               &$F_{1,12}$ & 3.174                   & 1.495                   & 0.875                   \\
			                      & $p$ & 0.100                   & 0.245                   & 0.368                   \\ \hline
			Teleoperation*Segment          &$F_{1,12}$  & 22.228                  & 1.879                   & 29.028                  \\
			                        & $p$ & \textbf{0.001}          & 0.196                   & \textbf{\textless0.001} \\ \hline
			Expertise*Teleoperation & $F_{1,12}$  & 7.859                   & 0.678                   & 0.156                   \\
			*Segment                        & $p$ & \textbf{0.016}          & 0.426                   & 0.700  \\       
			
			\botrule

	\end{tabular}}
	\raggedright{ 3-way mixed model ANOVA. Bold font indicates statistically significant effects.}
\end{tablehere}

\begin{figurehere}
	\begin{center}
		\centerline{\includegraphics{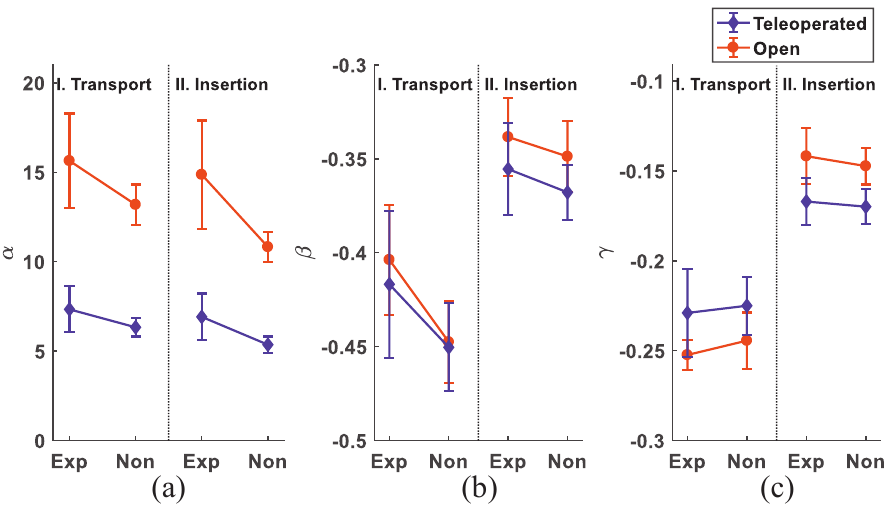}}
		\caption{The effects of teleoperatiom, experience, and movement segment on the power law parameters in the dVRK-Open dataset. The values of the speed gain factor $\alpha$ (a), the curvature exponent $\beta$ (b), and torsion exponent $\gamma$ (c) of experienced surgeons (Exp) and Non-medical users (Non) in the teleoperated and the open conditions for each segment. Markers are means, error bars are 95\% confidence intervals.}
		\label{fig5}
	\end{center}
\end{figurehere}
  by a distinctive color and marker. Our first question of interest focused on the global dependence of the power law parameters on the GRS. To appreciate these effects, we looked at all these data points together ignoring the information about the surgeon's identity. The summary of this analysis is the black dashed lines and the gray 95\% confidence interval areas. There was no statistically significant global effect of expertise level (as measured by the GRS) on the parameters $\alpha$ and $\gamma$. However, expertise level had 
\begin{tablehere}
	\tbl{Statistical Analysis Summary -- JIGSAWS \label{tab3}}
	{\begin{tabular}{@{}lcccr@{}}
			\toprule
			Model & Stat   & $\alpha$     & $\beta$                 & $\gamma$                 \\\hline\hline
			
			\multicolumn{5}{c}{Expertise} \\\hline\hline

Regression     & $t_{37}$  & 1.459          & -2.582         & -0.349         \\
             & $p$ & 0.153          & \textbf{0.014}          & 0.729          \\ \hline
 
   Repeated-measures & $F_{1,30}$  & 0.201          & 0.067          & 1.662          \\
              regression & $p$ & 0.658          & 0.798          & 0.207          \\ \hline\hline
              			\multicolumn{5}{c}{Segment} \\\hline\hline

               Repeated-measures         & $F_{3,21}$  & 37.284         & 32.603         & 31.230         \\
               ANOVA      & $p$ & \textbf{\textless0.001} & \textbf{\textless0.001} & \textbf{\textless0.001} \\

			\botrule

	\end{tabular}}
	\raggedright{Bold font indicates statistically significant effects.}
	%	Exp., Nov., and Tele. are abbreviations for experienced surgeons, novices, and teleoperation condition , respectively.}
	
\end{tablehere}

\begin{figurehere}
	\begin{center}
		\centerline{\includegraphics{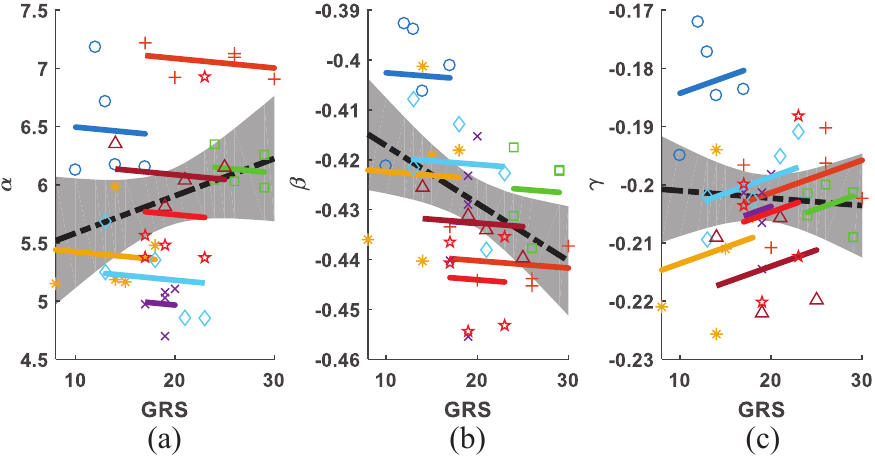}}
		\caption{The dependence of the power law parameters on the level of expertise as evaluated with the GRS in the JIGSAWS dataset. The values of the speed gain factor $\alpha$ (a), the curvature exponent $\beta$ (b), and torsion exponent $\gamma$ (c) of each trial are depicted as a function of expertise level (GRS). The different colors and markers indicate the eight participants. The black dashed lines are the linear regression lines, and the gray areas around them are 95\% confidence interval of the regression lines. The colored lines represent the fit of the repeated-measures regression model.}
		\label{fig6}
	\end{center}
\end{figurehere}
significant global effect on the parameter $\beta$: it decreased as the GRS increased. 

Our second question of interest was whether within the different trials of each surgeon, there is an effect of the GRS on the power law parameters. The summary of this analysis is highlighted in the colored lines in Fig.~\ref{fig6}. These lines were fitted with a repeated-measures regression model, which yields a single slope estimation, but each participant has a different intercept. The results of this analysis show that the GRS of each individual trial had no significant effect on the distribution of values of all the one-sixth power law parameters ($\alpha$, $\beta$, and $\gamma$). This together with the previous result suggests that the global dependency of $\beta$ on GRS was due to difference between the surgeons, and that the parameter $\beta$ can be useful for classification of surgical expertise, but not of the quality of the individual movement.

Our last question was how movement segment affects the parameters of the power law. Fig.~\ref{fig7} depicts the values of the power law in the four segments of the suturing task. A prominent observation from Fig.~\ref{fig7} is the big differences between the segments. The results of repeated-measures model ANOVA support this observation, and show that there were statistically significant differences between the segments for all the three parameters -- $\alpha$, $\beta$, and $\gamma$. These results suggest that the values of the power law may be useful for segmentation of complex surgical tasks.

\section{Discussion}
In this study, we characterized teleoperated and open suturing movements that was performed by participants with different levels of surgical skill. We analyzed the relationship between the speed of the suturing movements and their geometry -- path curvature and torsion -- in data from two different datasets. We found that the suturing movements that we analyzed were in accordance with the one-sixth power law. The fit of the power law was very good compared to the typical goodness of fit in such analyses suporting the assertion that indeed there is a characteristic relationship between the speed of the movements and their path curvature and torsion. In addition, we found that expertise level, teleoperation, and movement segment affect this relationship. In this section, we will discuss each of these effects.

\subsection{Surgical Expertise}
%expertise-alpha 
Our results suggest that the experience of a surgeon affects the parameters of the power law. This finding is in agreement with a previous study that reported that the speed gain factor $\alpha$ from a two-thirds power law can be useful for predicting surgical skill \cite{RefWorks:115}. In our analysis of the dVRK-Open dataset, we found the speed gain factor $\alpha$ of experienced surgeons was larger than of non-medical users. In contrast, in the JIGSAWS dataset, although it seems that $\alpha$ increases with the GRS (expertise level), this effect was not statistically significant. However, in the dVRK-Open dataset, the non-medical users had no experience in any kind of surgery, and they were all engineering students, and the experienced surgeons were all active robotic surgeons who performed at least 100 robotic cases. Therefore, even though surgical case experience does not necessarily reflect skill level, it is still very likely that the difference in the level of expertise between the non-medical users and the experienced surgeons was large. In contrast, in the JIGSAWS dataset, all the participants had surgical experience, and therefore, it is likely that the difference between their expertise level was relatively small. This difference between datasets may explain the difference in the results.

	 The speed gain factor $\alpha$ is related to overall tempo of motion \cite{RefWorks:100} (see Eq. (3)).
	 Hence, the effect of expertise level
	 \begin{figurehere}
	 	\begin{center}
	 		\centerline{\includegraphics{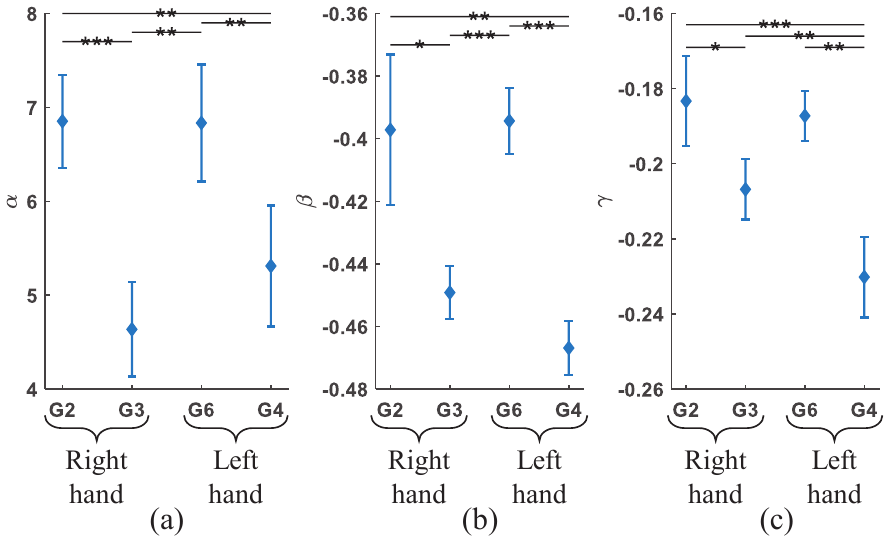}}
	 		\caption{The effect of movement segment on the power law parameters in the JIGSAWS dataset. The values of the speed gain factor $\alpha$ (a), the curvature exponent $\beta$ (b), and torsion exponent $\gamma$ (c) for each segment: G2. positioning, G3. pushing, G6. pulling, and G4. transferring. Markers are means, error bars are 95\% confidence intervals. *$p_B<0.05$, **$p_B<0.01$, ***$p_B<0.001$.}
	 		\label{fig7}
	 	\end{center}
	 \end{figurehere}
 on $\alpha$ may be related to the well-documented differences between completion times of participants with different surgical skill levels \cite{RefWorks:61, RefWorks:62, RefWorks:63}. In addition, the planar two-third power law can be predicted by smoothness of movement maximization  \cite{RefWorks:154}, and hence, the relation between expertise and the parameters of the power law may also be related to the smoothness of movement that is characteristic of skilled performance \cite{RefWorks:80,RefWorks:148}. 
	 
	 It is important to note that fitting $\alpha$ together with $\beta$ and $\gamma$ as we did in this study, leads to dependence of the units of $\alpha$ on the fitted value of $\beta$ and $\gamma$. It also leads to sensitivity of $\alpha$ to the units used for calculations. Hence, future work is needed to investigate whether a metric for surgical expertise level that will be based on the speed gain factor $\alpha$, as we calculated it, will be successful in classifying between expertise levels.

%expertise-beta
The value of the exponent $\beta$ in JIGSAWS decreased as the surgical skill (GRS) increased. This results suggests that $\beta$ may be used as an objective metric for surgical skill. However, this hypothesis should be explored in depth, with a greater number of movements and participants. For example, even in our study here, in the dVRK-Open dataset, the difference between $\beta$ values of experienced surgeons and non-medical users was not statistically significant, and seems to be reversed: experienced surgeons had higher $\beta$ than the non-medical users. A possible explanation for the discrepancy between the results of the two datasets is the differences in ages between the participants \cite{RefWorks:169}. The non-medical users in dVRK-Open were graduate students, and hence were much younger than the experienced surgeons. Thus, the results in dVRK-Open dataset may be due to the combined effect of surgical expertise and differences in age. To isolate the origin of the discrepancies, an experiment with age-matched non-medical users and experienced surgeons is needed.

\subsection{Teleoperation}
%teleoperation- alpha
The results from the dVRK-Open dataset show that using a robot-assisted surgery system to perform the suturing task significantly affected the one-sixth power law parameters. The difference between the speed gain factor $\alpha$ values from the teleoperated and the open sessions is likely due to the fact that we analyzed the data from the PSMs. To control the PSM, the speed of the MTM was down-scaled by a factor of 3. Because the speed gain factor $\alpha$ is related to overall tempo of motion  \cite{RefWorks:100}, the difference in $\alpha$ values between the conditions is likely a result of this down-scaling.
%teleoperation- beta
We hypothesized was that the  dynamic constraints in the teleoperation movements will affect the one-sixth power law parameters. Indeed, in the dVRK-Open dataset the curvature exponent $\beta$ differed between the teleoperated and the open conditions. In addition, the average $\beta$ values that obtained from the JIGSAWS dataset, in which all the movements were teleoperated, and the teleoperated condition in dVRK-Open dataset were similar. These results are consistent with the findings in \cite{RefWorks:116}, who suggested that dynamics can affect the relationship between the speed and curvature in planar movements. 

%teleoperation- gamma
 In dVRK-open, for the torsion exponent $\gamma$, the effect of teleoperation depended on the segment. This result may be due to the different nature of the two segments:  segment I (transport) was a movement in the air, whereas segment II (insertion) included interaction with tissue. In segment I, the  movement was not likely affected by the absence of haptic feedback in the teleoperated condition, and thus, in segment I the difference between teleoperated and open movements was relatively small. On the other hand, in segment II, there was interaction with tissue, the  movement was probably affected by haptic feedback, and therefore, the difference between teleoperated and open movement was greater. Further investigation of these findings is needed to examine whether this result was indeed caused by the haptic feedback.
 
  %teleoperation- conclusion
 Our results show that the use of surgical robot affect the parameters of the power law. This adds to the previously reported results of effects of robotic manipulators on movements of surgeons \cite{RefWorks:36,RefWorks:8}. Robotic assistance systems are developed today to help surgeons to perform and learn surgical tasks. For example, virtual fixtures are used to help surgeons to move the tools along specific path \cite{RefWorks:123,RefWorks:124,RefWorks:127, RefWorks:155}, or influencing the learning processes by applying forces that push the surgeon away from desired path or back to the desired path during training \cite{RefWorks:136}. In \cite{RefWorks:135}, researches showed that participants applied lower force when they held an end-effector that moved in accordance with the two-thirds power law comparing to an end-effector that moved not according to the power law. Their result suggests that for easy and intuitive assistance, it may be helpful to design shared control \cite{RefWorks:156} and assistive systems in a way that is compatible with the natural movement of humans. Using the one-sixth power law and other invariants from computational motor control may help in characterizing  natural movements, and improve the design of surgical robots. This can also be relevant to other applications of teleoperation \cite{RefWorks:157, RefWorks:158, RefWorks:159}, shared control \cite{RefWorks:160,RefWorks:161}, and devices such as exoskeletons \cite{RefWorks:162} and prostheses. Geometric invariance was even proposed to be useful principle for robotic trajectory generation \cite{RefWorks:163,RefWorks:164}.

\subsection{Movement Segment}
%segmentation
We found that all the power law parameters ($\alpha$, $\beta$, and $\gamma$) were affected by movement segment. These results were obtained from both datasets. In dVRK-open, the movements performed using one hand, and we investigated two segments. In JIGSAWS, we investigated four segments-- two segments performed using the right hand, and two using the left hand. Our results showed that for each power law parameter, the differences between segments that performed using the same hand  (I transport -- II insertion,  G2 positioning -- G3 pushing, and G6 pulling -- G4 transferring) were significant. These differences suggest that power law parameters may be used for development of automatic segmentation algorithms. In this study, we calculated the power law parameters on segments that were already set. Previous studies suggested that the speed gain factor $\alpha$ can be used for segmentation of surgical task \cite{RefWorks:120,RefWorks:121}. In future studies, we will develop and examine a segmentation algorithm that will divide surgical task into small segments based on more parameters of the power law ($\beta$, and $\gamma$). We will compare our algorithm to other algorithms for segmentation of surgical tasks \cite{RefWorks:85,RefWorks:87,RefWorks:105,RefWorks:170}.

There is a similarity between the segments from dVRK-open and JIGSAWS: I transport  and G2 positioning, II insertion and G3 pushing. Despite these similarities, it seems that for the parameters $\beta$ and $\gamma$, the difference between I transport and II insertion (Fig.~\ref{fig5}) is not in the same direction as the difference between G2 positioning and G3 pushing (Fig.~\ref{fig7}). However, a careful examination suggests that the values of $\beta$ and $\gamma$ that obtained from JIGSAWS, are all in a similar range as those parameters in segment I from dVRK-Open. As noted earlier, dynamic factors can affect the parameters of the power law \cite{RefWorks:116}. The tissues used in the experiments carried out to collect the two datasets were different. The tissue in dVRK-Open was made of foam, while in JIGSAWS, the tissue is made of an elastic fabric. Pushing the needle through the foam in dVRK-Open is very different from pushing a needle in the air, but pushing a needle through the elastic fabric may be more similar to pushing a needle in the air. Thus, in dVRK-Open, the values in segment II were higher than in segment I, and in JIGSAWS, the values in G3 were lower than in G2, but still within the range of the values of moment in the air. This explanation can be explored in future studies by analyzing movements that will interact with tissues of various properties. 

A segmentation algorithm that may be developed using the one-sixth power law parameters will highlight segmentation based on motor primitives \cite{RefWorks:19} rather than to procedural segments. Such kinematic segmentation may compliment the large body of literature on surgical task segmentation using contextual information \cite{RefWorks:85,RefWorks:87,RefWorks:105,RefWorks:170}. Combining these approaches with the motor primitives may lead to the development of even more successful algorithms for the important task of surgical movement segmentation. 

In computational motor control, there is a debate whether indeed segmented kinematics reflect segmented control strategies \cite{RefWorks:100,RefWorks:165}. However, there is increased body of evidence supporting segmented, or intermittent control \cite{RefWorks:113,RefWorks:88,RefWorks:166,RefWorks:167}, and movement chunking \cite{RefWorks:168}. Certainly, the debate about the neural origin of this segmentation does not impede the utility of our proposed analysis in understanding of surgical movements.

\section{Conclusion}
Our results demonstrate the potential benefit of characterizing surgical movement with an analysis of the relation between movement speed and its geometrical properties -- the one-sixth power law. We believe that this opens a promising avenue for improving the evaluation of surgical skill, teleoperation control design, and surgical gestures segmentation.

\nonumsection{Acknowledgments}
\noindent We thank Allison Okamura, Michael Hsieh, Zhan Fan Quek and Yuhang Che for their help in obtaining the experimental data. We thank Tamar Flash for useful discussions on the one-sixth power law. This study was supported in part by the Helmsley Charitable Trust through the Agricultural, Biological and Cognitive Robotics Initiative and by the Marcus Endowment Fund both at Ben-Gurion University of the Negev, and the ISF grant number 823/15. YS was supported by the Besor Fellowship.

\end{multicols}

\begin{thebibliography}{10}
	
	\bibitem{RefWorks:71}
	K.~A. Ericsson, Deliberate practice and the acquisition and maintenance of
	expert performance in medicine and related domains, {\em Acad. Med.} {\bf
		79}(10)  (2004)  S70--S81.
	
	\bibitem{RefWorks:138}
	R.~K. Reznick and H.~MacRae, Teaching surgical skills - changes in the wind,
	{\em New Engl. J. Med.} {\bf 355}(25)  (2006)  2664--2669.
	
	\bibitem{RefWorks:153}
	D.~M. Wolpert, Z.~Ghahramani and M.~I. Jordan, An internal model for
	sensorimotor integration, {\em Science} {\bf 269}(5232)  (1995)  1880--1882.
	
	\bibitem{RefWorks:113}
	A.~Karniel, Open questions in computational motor control, {\em J. Integr.
		Neurosci.} {\bf 10}(03)  (2011)  385--411.
	
	\bibitem{RefWorks:142}
	S.~Schaal and N.~Schweighofer, Computational motor control in humans and
	robots, {\em Curr. Opin. Neurobiol.} {\bf 15}(6)  (2005)  675--682.
	
	\bibitem{RefWorks:143}
	M.~Kawato, Internal models for motor control and trajectory planning, {\em
		Curr. Opin. Neurobiol.} {\bf 9}(6)  (1999)  718--727.
	
	\bibitem{RefWorks:144}
	J.~W. Krakauer and P.~Mazzoni, Human sensorimotor learning: adaptation, skill,
	and beyond, {\em Curr. Opin. Neurobiol.} {\bf 21}(4)  (2011)  636--644.
	
	\bibitem{RefWorks:145}
	R.~Shadmehr and S.~Mussa-Ivaldi, {\em Biological learning and control: how the
		brain builds representations, predicts events, and makes decisions} (Mit
	Press, Cambridge, MA, 2012).
	
	\bibitem{RefWorks:146}
	P.~Morasso, Spatial control of arm movements, {\em Exp. Brain. Res.} {\bf
		42}(2)  (1981)  223--227.
	
	\bibitem{RefWorks:39}
	T.~Flash and N.~Hogan, The coordination of arm movements: an experimentally
	confirmed mathematical model, {\em J. Neurosci.} {\bf 5}(7)  (1985)
	1688--1703.
	
	\bibitem{RefWorks:15}
	F.~Lacquaniti, C.~Terzuolo and P.~Viviani, The law relating the kinematic and
	figural aspects of drawing movements, {\em Acta Psychol.} {\bf 54}(1)  (1983)
	115--130.
	
	\bibitem{RefWorks:95}
	F.~E. Pollick, U.~Maoz, A.~A. Handzel, P.~J. Giblin, G.~Sapiro and T.~Flash,
	Three-dimensional arm movements at constant equi-affine speed, {\em Cortex}
	{\bf 45}(3)  (2009)  325--339.
	
	\bibitem{RefWorks:42}
	U.~Maoz, A.~Berthoz and T.~Flash, Complex unconstrained three-dimensional hand
	movement and constant equi-affine speed, {\em J. Neurophysiol.} {\bf 101}(2)
	(2009)  1002--1015.
	
	\bibitem{RefWorks:34}
	V.~Datta, S.~Mackay, M.~Mandalia and A.~Darzi, The use of electromagnetic
	motion tracking analysis to objectively measure open surgical skill in the
	laboratory-based model, {\em J. Am. Coll. Surg.} {\bf 193}(5)  (2001)
	479--485.
	
	\bibitem{RefWorks:75}
	M.~K. Chmarra, N.~H. Bakker, C.~A. Grimbergen and J.~Dankelman, Trendo, a
	device for tracking minimally invasive surgical instruments in training
	setups, {\em Sens. Actuators. A. Phys.} {\bf 126}(2)  (2006)  328--334.
	
	\bibitem{RefWorks:35}
	M.~S. Wilson, A.~Middlebrook, C.~Sutton, R.~Stone and R.~F. McCloy, Mist vr: a
	virtual reality trainer for laparoscopic surgery assesses performance, {\em
		Ann. R. Coll. Surg. Engl.} {\bf 79}(6)  (1997)  403--404.
	
	\bibitem{RefWorks:79}
	J.~Eriksen and T.~Grantcharov, Objective assessment of laparoscopic skills
	using a virtual reality stimulator, {\em Surg. Endosc.} {\bf 19}(9)  (2005)
	1216--1219.
	
	\bibitem{RefWorks:119}
	G.~S. Guthart and J.~K. Salisbury, The intuitive\textsuperscript{TM}
	telesurgery system: overview and application, {\em IEEE Int. Conf. Robot.
		Autom.\/},   (2000), pp. 618--621.
	
	\bibitem{RefWorks:33}
	P.~Kazanzidesf, Z.~Chen, A.~Deguet, G.~S. Fischer, R.~H. Taylor and S.~P.
	DiMaio, An open-source research kit for the da vinci® surgical system, {\em
		IEEE Int. Conf. Robot. Autom.\/},   (2014), pp. 6434--6439.
	
	\bibitem{RefWorks:137}
	A.~M. Jarc and I.~Nisky, Robot-assisted surgery: an emerging platform for human
	neuroscience research, {\em Front. Hum. Neurosci.} {\bf 9}  (2015).
	
	\bibitem{RefWorks:27}
	J.~D. Birkmeyer, J.~F. Finks, A.~O'Reilly, M.~Oerline, A.~M. Carlin, A.~R.
	Nunn, J.~Dimick, M.~Banerjee and N.~J. Birkmeyer, Surgical skill and
	complication rates after bariatric surgery, {\em N. Engl. J. Med.} {\bf
		369}(15)  (2013)  1434--1442.
	
	\bibitem{Refworks:147}
	M.~C. Vassiliou, L.~S. Feldman, C.~G. Andrew, S.~Bergman, K.~Leffondré,
	D.~Stanbridge and G.~M. Fried, A global assessment tool for evaluation of
	intraoperative laparoscopic skills, {\em Am. J. Surg.} {\bf 190}(1)  (2005)
	107--113.
	
	\bibitem{RefWorks:31}
	J.~Martin, G.~Regehr, R.~Reznick, H.~MacRae, J.~Murnaghan, C.~Hutchison and
	M.~Brown, Objective structured assessment of technical skill (osats) for
	surgical residents, {\em Br. J. Surg.} {\bf 84}(2)  (1997)  273--278.
	
	\bibitem{RefWorks:2}
	A.~V. Rij, J.~McDonald, R.~Pettigrew, M.~Putterill, C.~Reddy and J.~Wright,
	Cusum as an aid to early assessment of the surgical trainee, {\em Br. J.
		Surg.} {\bf 82}(11)  (1995)  1500--1503.
	
	\bibitem{RefWorks:62}
	T.~J. Tausch, T.~M. Kowalewski, L.~W. White, P.~S. McDonough, T.~C. Brand and
	T.~S. Lendvay, Content and construct validation of a robotic surgery
	curriculum using an electromagnetic instrument tracker, {\em J. Urol.} {\bf
		188}(3)  (2012)  919--923.
	
	\bibitem{RefWorks:63}
	N.~Rittenhouse, B.~Sharma, R.~Sonnadara, A.~Mihailidis and T.~Grantcharov,
	Design and validation of an assessment tool for open surgical procedures,
	{\em Surg. Endosc.} {\bf 28}(3)  (2014)  918--924.
	
	\bibitem{RefWorks:66}
	P.~A. Kenney, M.~F. Wszolek, J.~J. Gould, J.~A. Libertino and A.~Moinzadeh,
	Face, content, and construct validity of dv-trainer, a novel virtual reality
	simulator for robotic surgery, {\em Urology} {\bf 73}(6)  (2009)  1288--1292.
	
	\bibitem{RefWorks:61}
	E.~F. Hofstad, C.~Våpenstad, M.~K. Chmarra, T.~Langø, E.~Kuhry and
	R.~Mårvik, A study of psychomotor skills in minimally invasive surgery: what
	differentiates expert and nonexpert performance, {\em Surg. Endosc.} {\bf
		27}(3)  (2013)  854--863.
	
	\bibitem{RefWorks:56}
	K.~V. Sickle, D.~M. III, A.~Gallagher and C.~Smith, Construct validation of the
	promis simulator using a novel laparoscopic suturing task, {\em Surg.
		Endosc.} {\bf 19}(9)  (2005)  1227--1231.
	
	\bibitem{RefWorks:80}
	S.~Estrada, C.~Duran, D.~Schulz, J.~Bismuth, M.~D. Byrne and M.~K. O’Malley,
	Smoothness of surgical tool tip motion correlates to skill in endovascular
	tasks, {\em IEEE T. Hum.-Mach. Syst} {\bf 46}(5)  (2016)  647--659.
	
	\bibitem{RefWorks:148}
	K.~Liang, Y.~Xing, J.~Li, S.~Wang, A.~Li and J.~Li, Motion control skill
	assessment based on kinematic analysis of robotic end-effector movements,
	{\em Int. J. Med. Robot.}   (2017) p. e1845.
	
	\bibitem{RefWorks:149}
	B.~Poursartip, M.-E. LeBel, L.~C. McCracken, A.~Escoto, R.~V. Patel, M.~D.
	Naish and A.~L. Trejos, Energy-based metrics for arthroscopic skills
	assessment, {\em Sensors} {\bf 17}(8)  (2017) p. 1808.
	
	\bibitem{RefWorks:150}
	B.~Poursartip, M.~E. LeBel, R.~Patel, M.~Naish and A.~L. Trejos, Analysis of
	energy-based metrics for laparoscopic skills assessment, {\em IEEE Trans.
		Biomed. Eng.} {\bf PP}(99)  (2017).
	
	\bibitem{RefWorks:151}
	J.~D. Brown, C.~E. O’Brien, S.~C. Leung, K.~R. Dumon, D.~I. Lee and K.~J.
	Kuchenbecker, Using contact forces and robot arm accelerations to
	automatically rate surgeon skill at peg transfer, {\em IEEE Trans. Biomed.
		Eng.} {\bf 64}(9)  (2017)  2263--2275.
	
	\bibitem{RefWorks:5}
	S.~Maeso, M.~Reza, J.~A. Mayol, J.~A. Blasco, M.~Guerra, E.~Andradas and M.~N.
	Plana, Efficacy of the da vinci surgical system in abdominal surgery compared
	with that of laparoscopy: a systematic review and meta-analysis, {\em Ann.
		Surg.} {\bf 252}(2)  (2010)  254--262.
	
	\bibitem{RefWorks:50}
	K.~Moorthy, Y.~Munz, A.~Dosis, J.~Hernandez, S.~Martin, F.~Bello, T.~Rockall
	and A.~Darzi, Dexterity enhancement with robotic surgery, {\em Surg. Endosc.}
	{\bf 18}(5)  (2004)  790--795.
	
	\bibitem{RefWorks:72}
	A.~R. Lanfranco, A.~E. Castellanos, J.~P. Desai and W.~C. Meyers, Robotic
	surgery: a current perspective, {\em Ann. Surg.} {\bf 239}(1)  (2004)
	14--21.
	
	\bibitem{RefWorks:73}
	A.~M. Okamura, Haptic feedback in robot-assisted minimally invasive surgery,
	{\em Curr. Opin. Urol.} {\bf 19}(1)  (2009)  102--107.
	
	\bibitem{RefWorks:8}
	I.~Nisky, M.~H. Hsieh and A.~M. Okamura, Uncontrolled manifold analysis of arm
	joint angle variability during robotic teleoperation and freehand movement of
	surgeons and novices, {\em IEEE Trans. Biomed. Eng.} {\bf 61}(12)  (2014)
	2869--2881.
	
	\bibitem{RefWorks:36}
	I.~Nisky, A.~M. Okamura and M.~H. Hsieh, Effects of robotic manipulators on
	movements of novices and surgeons, {\em Surg. Endosc.} {\bf 28}(7)  (2014)
	2145--2158.
	
	\bibitem{RefWorks:82}
	H.~C. Lin, I.~Shafran, D.~Yuh and G.~D. Hager, Towards automatic skill
	evaluation: Detection and segmentation of robot-assisted surgical motions,
	{\em Comput. Aided Surg.} {\bf 11}(5)  (2006)  220--230.
	
	\bibitem{RefWorks:85}
	A.~James, D.~Vieira, B.~Lo, A.~Darzi and G.-Z. Yang, Eye-gaze driven surgical
	workflow segmentation, {\em Med. Image. Comput. Comput. Assist. Interv.\/},
	(2007), pp. 110--117.
	
	\bibitem{RefWorks:87}
	L.~Zappella, B.~Béjar, G.~Hager and R.~Vidal, Surgical gesture classification
	from video and kinematic data, {\em Med. Image Anal.} {\bf 17}(7)  (2013)
	732--745.
	
	\bibitem{RefWorks:105}
	N.~Ahmidi, L.~Tao, S.~Sefati, Y.~Gao, C.~Lea, B.~B. Haro, L.~Zappella,
	S.~Khudanpur, R.~Vidal and G.~D. Hager, A dataset and benchmarks for
	segmentation and recognition of gestures in robotic surgery, {\em IEEE Trans.
		Biomed. Eng.} {\bf 64}(9)  (2017)  2025--2041.
	
	\bibitem{RefWorks:170}
	F.~Despinoy, D.~Bouget, G.~Forestier, C.~Penet, N.~Zemiti, P.~Poignet and
	P.~Jannin, Unsupervised trajectory segmentation for surgical gesture
	recognition in robotic training, {\em IEEE Trans. Biomed. Eng.} {\bf 63}(6)
	(2016)  1280--1291.
	
	\bibitem{RefWorks:19}
	P.~Viviani and M.~Cenzato, Segmentation and coupling in complex movements.,
	{\em J. Exp. Psychol.-Hum. Percept. Perform.} {\bf 11}(6)  (1985) p. 828.
	
	\bibitem{RefWorks:11}
	I.~Nisky, Y.~Che, Z.~F. Quek, M.~Weber, M.~H. Hsieh and A.~M. Okamura,
	Teleoperated versus open needle driving: Kinematic analysis of experienced
	surgeons and novice users, {\em IEEE Int. Conf. Robot. Autom.\/},   (2015),
	pp. 5371--5377.
	
	\bibitem{RefWorks:112}
	Y.~Gao, S.~S. Vedula, C.~E. Reiley, N.~Ahmidi, B.~Varadarajan, H.~C. Lin,
	L.~Tao, L.~Zappella, B.~Béjar and D.~D. Yuh, The jhu-isi gesture and skill
	assessment working set (jigsaws): A surgical activity dataset for human
	motion modeling, {\em Model. Monit. Comput. Assist. Interv.\/},   {\bf 3}
	(2014).
	
	\bibitem{RefWorks:97}
	F.~E. Pollick and G.~Sapiro, Constant affine velocity predicts the 1/3 power
	law of planar motion perception and generation, {\em Vision Res.} {\bf 37}(3)
	(1997)  347--353.
	
	\bibitem{RefWorks:98}
	T.~Flash and A.~A. Handzel, Affine differential geometry analysis of human arm
	movements, {\em Biol. Cybern.} {\bf 96}(6)  (2007)  577--601.
	
	\bibitem{RefWorks:17}
	P.~Viviani and R.~Schneider, A developmental study of the relationship between
	geometry and kinematics in drawing movements., {\em J. Exp. Psychol.-Hum.
		Percept. Perform.} {\bf 17}(1)  (1991) p. 198.
	
	\bibitem{RefWorks:21}
	C.~de'Sperati and P.~Viviani, The relationship between curvature and velocity
	in two-dimensional smooth pursuit eye movements, {\em J. Neurosci.} {\bf
		17}(10)  (1997)  3932--3945.
	
	\bibitem{RefWorks:20}
	S.~Vieilledent, Y.~Kerlirzin, S.~Dalbera and A.~Berthoz, Relationship between
	velocity and curvature of a human locomotor trajectory, {\em Neurosci. Lett.}
	{\bf 305}(1)  (2001)  65--69.
	
	\bibitem{RefWorks:152}
	M.~Karklinsky and T.~Flash, Timing of continuous motor imagery: the two-thirds
	power law originates in trajectory planning, {\em J. Neurophysiol.} {\bf
		113}(7)  (2015)  2490--2499.
	
	\bibitem{RefWorks:96}
	S.~Schaal and D.~Sternad, Origins and violations of the 2/3 power law in
	rhythmic three-dimensional arm movements, {\em Exp. Brain Res.} {\bf 136}(1)
	(2001)  60--72.
	
	\bibitem{RefWorks:115}
	A.~French, T.~S. Lendvay, R.~M. Sweet and T.~M. Kowalewski, Predicting surgical
	skill from the first n seconds of a task: value over task time using the
	isogony principle, {\em Int. J. Comput. Assist. Radiol. Surg.} {\bf 12}(7)
	(2017)  1161--1170.
	
	\bibitem{RefWorks:116}
	G.~Catavitello, Y.~P. Ivanenko, F.~Lacquaniti and P.~Viviani, Drawing ellipses
	in water: evidence for dynamic constraints in the relation between velocity
	and path curvature, {\em Exp. Brain Res.} {\bf 234}(6)  (2016)  1649--1657.
	
	\bibitem{RefWorks:117}
	D.~Huh and T.~J. Sejnowski, Spectrum of power laws for curved hand movements,
	{\em Proc. Natl. Acad. Sci. U. S. A.} {\bf 112}(29)  (2015)  E3950--E3958.
	
	\bibitem{RefWorks:99}
	D.~M. Endres, Y.~Meirovitch, T.~Flash and M.~A. Giese, Segmenting sign language
	into motor primitives with bayesian binning, {\em Front. Comput. Neurosci.}
	{\bf 7}  (2013) p.~68.
	
	\bibitem{RefWorks:120}
	J.~Cifuentes, P.~Boulanger, M.~T. Pham, R.~Moreau and F.~Prieto, Automatic
	gesture analysis using constant affine velocity, {\em Conf. Proc. IEEE Eng.
		Med. Biol. Soc.\/},   (2014), pp. 1826--9.
	
	\bibitem{RefWorks:121}
	S.~B. Shafiei, K.~A. Guru and E.~T. Esfahani, Using two-third power law for
	segmentation of hand movement in robotic assisted surgery, {\em IDETC/CIE
		ASME\/},   (2015).
	
	\bibitem{RefWorks:171}
	Y.~Sharon, T.~S. Lendvay and I.~Nisky, Instrument orientation-based metrics for
	surgical skill evaluation in robot-assisted and open needle driving, {\em
		arXiv preprint arXiv:1709.09452}   (2017).
	
	\bibitem{RefWorks:100}
	M.~J. Richardson and T.~Flash, Comparing smooth arm movements with the
	two-thirds power law and the related segmented-control hypothesis, {\em J.
		Neurosci.} {\bf 22}(18)  (2002)  8201--8211.
	
	\bibitem{RefWorks:154}
	T.~Flash, Y.~Meirovitch and A.~Barliya, Models of human movement: Trajectory
	planning and inverse kinematics studies, {\em Robot. Auton. Syst.} {\bf
		61}(4)  (2013)  330--339.
	
	\bibitem{RefWorks:169}
	T.~L. Gibo, A.~J. Bastian and A.~M. Okamura, Effect of age on stiffness
	modulation during postural maintenance of the arm, {\em IEEE Int. Conf.
		Rehabil. Robot.\/},   (2013), pp. 1--6.
	
	\bibitem{RefWorks:123}
	J.~Abbott, P.~Marayong and A.~Okamura, Haptic virtual fixtures for
	robot-assisted manipulation, {\em Int. Symp. Robot. Res.\/},   {\bf 28}
	(2007), pp. 49--64.
	
	\bibitem{RefWorks:124}
	S.~A. Bowyer, B.~L. Davies and Y.~B. Rodriguez, Active constraints/virtual
	fixtures: A survey, {\em IEEE Trans. Robot.} {\bf 30}(1)  (2014)  138--157.
	
	\bibitem{RefWorks:127}
	N.~Enayati, E.~D. Momi and G.~Ferrigno, Haptics in robot-assisted surgery:
	Challenges and benefits, {\em IEEE Rev. Biomed. Eng.} {\bf 9}  (2016)
	49--65.
	
	\bibitem{RefWorks:155}
	M.~A. Vitrani, C.~Poquet and G.~Morel, Applying virtual fixtures to the distal
	end of a minimally invasive surgery instrument, {\em IEEE Trans. Robot.} {\bf
		33}(1)  (2017)  114--123.
	
	\bibitem{RefWorks:136}
	M.~M. Coad, A.~M. Okamura, S.~Wren, Y.~Mintz, T.~S. Lendvay, A.~M. Jarc and
	I.~Nisky, Training in divergent and convergent force fields during 6-dof
	teleoperation with a robot-assisted surgical system, {\em IEEE World Haptic.
		Conf.\/},   (2017), pp. 195--200.
	
	\bibitem{RefWorks:135}
	P.~Maurice, M.~E. Huber, N.~Hogan and D.~Sternad, Velocity-curvature patterns
	limit human–robot physical interaction, {\em IEEE Robot. Autom. Lett.} {\bf
		3}(1)  (2018)  249--256.
	
	\bibitem{RefWorks:156}
	D.~A. Abbink, M.~Mulder and E.~R. Boer, Haptic shared control: smoothly
	shifting control authority?, {\em Cogn. Technol. Work} {\bf 14}(1)  (2012)
	19--28.
	
	\bibitem{RefWorks:157}
	N.~Jarrassé, J.~Paik, V.~Pasqui and G.~Morel, How can human motion prediction
	increase transparency?, {\em IEEE Int. Conf. Robot. Autom.\/},   (2008), pp.
	2134--2139.
	
	\bibitem{RefWorks:158}
	I.~Nisky, F.~A. Mussa-Ivaldi and A.~Karniel, Analytical study of perceptual and
	motor transparency in bilateral teleoperation, {\em IEEE T. Hum.-Mach. Syst}
	{\bf 43}(6)  (2013)  570--582.
	
	\bibitem{RefWorks:159}
	J.~Buzzi, C.~Gatti, G.~Ferrigno and E.~D. Momi, Analysis of joint and hand
	impedance during teleoperation and free-hand task execution, {\em IEEE Robot.
		Autom. Lett.} {\bf 2}(3)  (2017)  1733--1739.
	
	\bibitem{RefWorks:160}
	H.~Boessenkool, D.~A. Abbink, C.~J.~M. Heemskerk, F.~C.~T. van~der Helm and
	J.~G.~W. Wildenbeest, A task-specific analysis of the benefit of haptic
	shared control during telemanipulation, {\em IEEE Trans. Haptics.} {\bf 6}(1)
	(2013)  2--12.
	
	\bibitem{RefWorks:161}
	J.~Smisek, E.~Sunil, M.~M. van Paassen, D.~A. Abbink and M.~Mulder,
	Neuromuscular-system-based tuning of a haptic shared control interface for
	uav teleoperation, {\em IEEE T. Hum.-Mach. Syst} {\bf 47}(4)  (2017)
	449--461.
	
	\bibitem{RefWorks:162}
	N.~Jarrasse and G.~Morel, Connecting a human limb to an exoskeleton, {\em IEEE
		Trans. Robot.} {\bf 28}(3)  (2012)  697--709.
	
	\bibitem{RefWorks:163}
	Y.~Meirovitch, D.~Bennequin and T.~Flash, Geometrical invariance and smoothness
	maximization for task-space movement generation, {\em IEEE Trans. Robot.}
	{\bf 32}(4)  (2016)  837--853.
	
	\bibitem{RefWorks:164}
	M.~Karklinsky, M.~Naveau, A.~Mukovskiy, O.~Stasse, T.~Flash and P.~Soueres,
	Robust human-inspired power law trajectories for humanoid hrp-2 robot, {\em
		IEEE Conf. Biomed. Robot. Biomechatronics\/},   (2016), pp. 106--113.
	
	\bibitem{RefWorks:165}
	D.~Sternad and S.~Schaal, Segmentation of endpoint trajectories does not imply
	segmented control, {\em Exp. Brain. Res.} {\bf 124}(1)  (1999)  118--136.
	
	\bibitem{RefWorks:88}
	R.~Leib and A.~Karniel, Minimum acceleration with constraints of center of
	mass: a unified model for arm movements and object manipulation, {\em J.
		Neurophysiol.} {\bf 108}(6)  (2012)  1646--1655.
	
	\bibitem{RefWorks:166}
	S.~Ben-Itzhak and A.~Karniel, Minimum acceleration criterion with constraints
	implies bang-bang control as an underlying principle for optimal trajectories
	of arm reaching movements, {\em Neural Comput.} {\bf 20}(3)  (2008)
	779--812.
	
	\bibitem{RefWorks:167}
	P.~Gawthrop, I.~Loram, M.~Lakie and H.~Gollee, Intermittent control: a
	computational theory of human control, {\em Biol. Cybern.} {\bf 104}(1)
	(2011)  31--51.
	
	\bibitem{RefWorks:168}
	P.~Ramkumar, D.~E. Acuna, M.~Berniker, S.~T. Grafton, R.~S. Turner and K.~P.
	Kording, Chunking as the result of an efficiency computation trade-off, {\em
		Nat. Commun.} {\bf 7}  (2016).
	
\end{thebibliography}
\end{document}